\documentclass[11pt,a4paper]{article}
\usepackage[margin=1in]{geometry}
\usepackage[T1]{fontenc}
\usepackage{lmodern,microtype,amsmath,amssymb,booktabs,longtable,array,graphicx,xcolor,float}
\usepackage[hidelinks]{hyperref}
\usepackage{caption}
\usepackage{setspace,ragged2e,xurl}
\usepackage{fancyhdr}
\title{Spatial disparities in simulated patrol detection:\\a reproducible generative-model audit across Baltimore and Chicago}
\author{Pronob Kumar Barman$^{1}$ \and Pronoy Kumar Barman$^{2}$}
\date{}
\hypersetup{pdftitle={Spatial disparities in simulated patrol detection: a reproducible generative-model audit across Baltimore and Chicago},pdfauthor={Pronob Kumar Barman and Pronoy Kumar Barman},pdfsubject={Revised arXiv preprint},pdfkeywords={predictive policing, spatial disparities, simulation, generative models}}
\begin{document}
\singlespacing
\raggedright
\hyphenpenalty=10000
\exhyphenpenalty=10000
\begingroup
\centering
{\singlespacing
\maketitle}
\begin{center}\small
$^{1}$Department of Information Systems, University of Maryland, Baltimore County (UMBC), Baltimore, Maryland, USA\\
$^{2}$Department of Statistics, Jagannath University, Dhaka, Bangladesh\\[4pt]
Corresponding author: Pronob Kumar Barman, \href{mailto:pbarman1@umbc.edu}{pbarman1@umbc.edu}
\end{center}
\endgroup\clearpage
\raggedright

\begin{abstract}
\textbf{Background.} Auditing predictive policing requires distinguishing recorded incidents, simulated police observations, and the demographic characteristics of places. We revise a spatial simulation framework that couples a generative adversarial network with a probabilistic patrol-detection model and evaluate it against empirical-location sampling and uniform-area deployment.

\textbf{Results.} A frozen archive contains 145178 Baltimore records from 2017--2019 and 234882 Chicago records from 2022. We rebuild geographic covariates from official Census tract boundaries and American Community Survey tables. Twenty independent seeds per condition produce 560 paired-condition trajectories, each spanning February--December. Under detection-only training, generative-policy mean Black-area minus White-area detection differences range from -2.03 to 0.75 percentage points across city-years, with coverage from 1.78\% to 7.50\%. Matched observation-stream contrasts assess feedback conditional on the recorded incident archive. A conditional tabular generative-model replacement experiment also tests whether synthetic labels agree with the demographics of generated locations. The specified synthesis procedure fails geographic balancing in 107 of 220 seed-month fits, leaving no complete annual synthetic-replacement trajectories.

\textbf{Conclusions.} Spatial detection disparities depend on the observation process, allocation model, training budget, and stochastic realization. The framework supports reproducible scenario comparison, but does not measure individual racial disparities in actual police contact or identify the distribution of unrecorded crime. Absolute detection differences, coverage, synthesis feasibility, and simulation uncertainty should accompany ratio-based fairness summaries.
\end{abstract}

\clearpage
\section{Background}
Predictive policing raises a measurement problem as well as a modeling problem. An administrative incident record reflects an event that entered an institutional recording system. A model trained on such records learns a distribution of recorded locations; it does not recover the complete spatial distribution of crime. If later training inputs depend on locations that officers observe, deployment and observation can become mutually dependent. Ensign et al. formalized this observation-feedback mechanism and showed how discovered and reported incidents interact within a predictive-policing model~\cite{ensign2018runaway}. Lum and Isaac's critique motivates examining the institutional consequences of learning from police data~\cite{lum2016predict}.

Historical data quality also affects what an audit can establish. Richardson et al. describe how civil-rights violations and police practices can contaminate administrative data~\cite{richardson2019dirty}. Treating citizen reporting as unqualified ground truth would bypass this concern. Under-reporting-aware prediction research instead explicitly models incomplete observation and examines resulting accuracy--fairness tradeoffs~\cite{wu2024fairness}. A simulation can expose consequences of an assumed observation mechanism, but it cannot validate that mechanism merely by producing numerical disparity estimates.

City-specific comparisons are therefore valuable, while claims of universal direction or magnitude require caution. Semsar et al. compare predictive and hotspot-policing simulations in Baltimore and show that fairness and accuracy can evolve differently over time~\cite{semsar2026comparative}. The present study does not claim to be the first simulation of policing feedback. Its contribution is an auditable comparison involving a generative spatial allocator, simpler allocation rules, explicitly defined observation streams, and repeated stochastic trajectories in two distinct urban datasets. The comparison evaluates a specified prototype rather than a deployed policing system or a state-of-the-art forecasting benchmark.

Generative adversarial networks learn to produce samples resembling training observations~\cite{goodfellow2014generative}. Conditional tabular generation extends this approach to mixed continuous and categorical data~\cite{xu2019modeling}; its ability to generate a category label does not itself establish that the label agrees with the geography of generated coordinates. More broadly, quantitative fairness criteria require a sociotechnical interpretation~\cite{selbst2019fairness}. In this study, higher detection is neither automatically beneficial nor automatically harmful: simulated detection does not measure protection, coercion, wrongful contact, deterrence, or community preferences.

\subsection{Related evidence and the scope of the audit}
Operational evaluations and simulation audits answer different questions. Mohler et al. conducted randomized controlled field trials of predictive policing, directly examining deployment under field conditions~\cite{mohler2015randomized}. Our experiment instead compares outcomes generated by a specified observation model. Ferguson's analysis of predictive-policing governance motivates evaluating how such tools are scrutinized, rather than treating a statistical disparity measure as a complete accountability framework~\cite{ferguson2017policing}.

Fairness criteria also describe different targets. Mehrabi et al. survey sources of bias and organize alternative fairness definitions~\cite{mehrabi2021survey}; Berk et al. examine tradeoffs among fairness and accuracy in criminal-justice risk assessments~\cite{berk2021fairness}. Chouldechova demonstrates incompatibilities among particular prediction criteria when outcome prevalence differs across groups~\cite{chouldechova2017fair}. That result is not a universal impossibility claim for every spatial-allocation objective. Our detection-rate gap, coverage and area-level Gini are descriptive outcomes of the specified simulation, rather than interchangeable definitions of justice.

Studies of recidivism additionally motivate explicit benchmarking against simpler methods. Dressel and Farid compare a commercial risk score with human judgments and a simple predictor~\cite{dressel2018accuracy}, while Wang et al. investigate interpretable recidivism models and their accuracy and fairness~\cite{wang2023interpretable}. Those studies predict outcomes for people; they do not establish racial detection rates or patrol effects in Baltimore or Chicago. We draw on their benchmarking rationale without treating their findings as evidence about our spatial outcome.

Causal and counterfactual fairness require further assumptions. Zhang and Bareinboim formulate explanations of disparity through structural causal mechanisms~\cite{zhang2018fairness}. Kim et al. develop a disentangled causal-effect variational autoencoder for counterfactual generation and fairness~\cite{kim2021counterfactual}. Our paired simulation contrasts manipulate observation streams within a model; they neither implement those causal identification strategies nor estimate an individual counterfactual effect of race.

Finally, the predictive-policing literature extends beyond one metric or protected attribute. Almasoud and Idowu review fairness strategies and identify gaps concerning protected attributes beyond race~\cite{almasoud2025bias}. Hung and Yen examine the social context and governance of algorithmic fairness~\cite{hung2023ethics}. Ziosi and Pruss document how stakeholders interpret and use evidence of algorithmic bias differently~\cite{ziosi2024contested}. These perspectives support a constrained interpretation of our results: a modeled detection difference does not alone establish public benefit, harm, legitimacy or an appropriate policy response.

We ask three questions. First, how do simulated detection differences and total coverage vary across allocation rules and observation streams? Second, how much does the observation-feedback mechanism change those outcomes relative to a matched full-record reference? Third, can a specified synthetic replacement procedure produce geographically consistent samples and improve disparity without concealing changes in coverage? Tract-level demographic associations are treated as an additional exploratory description rather than evidence of individual or causal racial discrimination.

\section{Methods}
\subsection{Design, data scope, and temporal ordering}
The analysis uses the four processed incident files supplied with the replication project. They contain 51511, 47856, and 45811 Baltimore records for 2017, 2018, and 2019, respectively, and 234882 Chicago records for 2022. These are counts of archived records, not estimates of all crimes. The source portals are the Baltimore legacy incident-data service and the Chicago Crimes dataset~\cite{BPD2019,CPD2022}. The original raw extracts, complete filtering logs, and original extraction dates were unavailable. We consequently identify and freeze the supplied processed files with cryptographic hashes rather than claim to reconstruct the original extraction. A single current Chicago record was spot-checked against the source and agreed; that check does not establish complete archive validity.

We retain the archive's incident universe, remove locations outside the fixed municipal analysis mask, and report the resulting counts in Table~\ref{tab:data}. The Baltimore mask is the union of its downloaded 2019 tracts; the Chicago mask comes from the city's published municipal boundary~\cite{chicagoBoundary}. The downloaded boundary versions are fixed for all comparisons. We do not assume that their administrative or geocoding precision exactly matches every historical incident. Baltimore and Chicago also differ in calendar period and offense coverage. Comparisons within each city-year hold the incident archive fixed; cross-city contrasts are descriptive and cannot isolate a city effect.

Each city-year is a separate January-initialized experiment. All January recorded incidents are available for initialization. February deployment uses January training observations; March uses the previous February observations; this continues through December. Current-month outcomes never enter that month's training. The three Baltimore years are independent annual sequences, not a continuous three-year feedback trajectory. A January initialization on all records is an informational assumption and is not an empirically verified account of what officers knew.

\subsection{Geographic covariates and the estimand}
We spatially join incident coordinates to Census tracts, then attach 2019 American Community Survey (ACS) five-year estimates for Baltimore and 2022 estimates for Chicago~\cite{ACS2022}. Tract geometries are obtained from the Census Bureau geography services and TIGER/Line files~\cite{censusGeography}. The Baltimore race table is B02001, income table B19013, and poverty table B17001; the same tables are used for Chicago. Black and White shares use the Black-alone and White-alone estimates divided by the total population. These categories can include Hispanic residents; ethnicity is not independently modeled. Poverty uses the population for whom poverty status is determined as its denominator. Table extracts, source identifiers, geographic files and the download manifest accompany the replication package.

A tract is classified as Black-majority if its Black-alone share exceeds 50\%, White-majority if its White-alone share exceeds 50\%, and Other otherwise. Other is a heterogeneous geographic category, not an individual racial identity. Zero-population, missing-demographic and unmatched areas are Unknown. No city-average demographic values are imputed. Unknown-area incidents remain eligible for training and total-coverage calculations but do not enter Black-versus-White rate denominators. Boundary ties are resolved deterministically by the smallest tract identifier. The fixed 2019 Baltimore covariates provide a retrospective geographic classification for all three years, not a real-time demographic forecast. ACS estimation uncertainty and classification uncertainty near 50\% are not propagated into the simulation intervals.

The primary group rate is the fraction of \emph{recorded incident locations} in an area category that receive simulated patrol detection. The numerator is a simulated detected-incident count, and the denominator is the archive's recorded-incident count for that area category and period. No victim, offender, or contacted person's race is inferred or sampled. Rates are not per capita and are not observed racial differences in police encounters.

\subsection{Allocation and observation processes}
Coordinates are transformed into local projected meters: EPSG:26985 for Baltimore and EPSG:26971 for Chicago. Every month a policy produces 60 patrol locations. For recorded incident $i$, let $k_i$ be the number of patrol locations within $700$ feet ($213.36$ m). The detection probability is
\begin{equation}
 p_i=1-(1-0.85)^{k_i},\qquad D_i\sim\operatorname{Bernoulli}(p_i).
\end{equation}
An independent citizen-report indicator is $C_i\sim\operatorname{Bernoulli}(0.521)$, and mixed observation is $M_i=D_i\lor C_i$. Thus the mixed stream is a union of patrol detections and citizen reports, not exclusively calls for service. The reporting parameter 0.521 is retained as a scenario assumption for comparability with the prototype; it is not presented as an empirically established reporting probability for these cities or offense mixtures. National victimization evidence shows that reporting varies by offense and period~\cite{BJS2022}; it does not justify a universal probability for the archived records.

The three training streams are: (i) \emph{detected}, using only the previous month's simulated patrol detections; (ii) \emph{mixed}, using the previous month's union of detections and citizen reports; and (iii) \emph{full-record reference}, using all previous-month archived incidents. The reference removes dependence of training inputs on simulated detection; it is not a latent-crime ground truth. If an observed training set contains fewer than ten records, the most recent eligible observed training set is carried forward and the event is logged. Unobserved previous-month records are never silently restored to the detected stream.

This is a static monthly exposure model: the same set of patrol locations is evaluated against all incidents in a month. It contains no patrol routes, shifts, travel times, response capacity, crime deterrence, displacement, arrests, or subsequent criminal-justice outcomes. Its 60 locations should not be interpreted as an operational staffing model.

\subsection{Generative model and simpler allocation rules}
The generator maps 100 independent standard-normal latent components through fully connected layers of sizes 256, 512, 256, and 2. The hidden layers use batch normalization and LeakyReLU with slope 0.2; the output uses tanh. The discriminator has hidden sizes 512, 256, and 128, LeakyReLU activations, dropout 0.3 after the first two layers, and a single logit. Real and generated coordinates share a fixed city-bounding-box transformation to $[-11]^2$. This replaces the prototype's mismatch between unbounded standardized real coordinates and tanh-bounded generated coordinates.

Training uses Adam with learning rate 0.0002 and coefficients $(0.50.999)$. Each monthly fit performs 200 optimizer updates, each using at most 64 real records sampled with replacement. These are updates, not 200 complete passes through the data. Binary cross-entropy with logits implements the discriminator objective and the non-saturating generator objective. The generator is put in evaluation mode before sampling. Generated coordinates outside the city mask are rejected; proposal rejection fractions are logged. A finite sampling limit makes deployment failure observable rather than substituting another policy without disclosure. The 200-update budget is specified for reproducibility and is not a convergence claim; a 1000-update sensitivity experiment examines budget dependence.

The \emph{empirical} baseline samples previous-month training coordinates with replacement, without fitting a generative model. It uses the same detected, mixed and reference streams. The \emph{uniform-area} baseline samples points uniformly within the city polygon. It does not learn from observations, so only one reference-labeled condition is needed. Uniform area includes nonresidential areas and is an allocation benchmark, not a recommendation about appropriate police presence. All policies use the same officer count, physical detection model, incident archive and paired random numbers for event detection and reporting.

\subsection{Outcomes, aggregation and uncertainty}
For group $g$, seed $s$ and an annual sequence of months $T$, the pooled detection rate is
\begin{equation}
 r_{gs}=\frac{\sum_{t\in T}N^{\mathrm{det}}_{gts}}{\sum_{t\in T}N^{\mathrm{total}}_{gt}}.
\end{equation}
The primary signed difference is $\Delta_s=100(r_{Bs}-r_{Ws})$ in percentage points (pp). We also report total detected records divided by all recorded incidents, including Unknown areas. A positive gap denotes higher modeled detection in Black-majority tracts; it does not imply a benefit or an identified harm. The secondary rate ratio is $r_B/r_W$. A positive numerator with a zero denominator produces infinity; zero/zero and absent-group cases are undefined. Counts of such cases accompany finite summaries. We do not replace zero rates by an arbitrary epsilon. Annual pooled ratios, ratios of mean seed-level rates and averages of monthly ratios are different estimands and are labeled accordingly.

The three-group unweighted Gini is retained as a secondary descriptive measure. Its unnormalized maximum is $2/3$ for three nonnegative rates. It is undefined when a group is absent or all rates are zero, and is not population-weighted inequality. The prototype's composite ``Bias Amplification Score'' is omitted: a composite disparity statistic alone does not identify a feedback effect.

Twenty independent seeds (100--119) are used per condition. There are seven conditions per city-year: three observation streams for each of the generative and empirical policies, plus uniform area. This yields 560 complete trajectories and 6160 dependent monthly policy evaluations. Conditions with the same seed are paired; the 560 trajectories are not all mutually independent replications. The pilot seed 0 is excluded. Four thousand nonparametric bootstrap resamples of whole seed-level trajectories provide 95\% intervals for mean outcomes. Monthly charts use 2000 whole-trajectory resamples for pointwise bands. Monte Carlo standard errors are reported, with mean-gap standard errors above 0.5 pp flagged as imprecise under the fixed 20-seed budget.

For feedback comparisons we compute paired differences relative to the same policy's reference trajectory. In addition to signed gap and coverage contrasts, the magnitude contrast is
\begin{equation}
 A_s=|\Delta_{s,\mathrm{observed}}|-|\Delta_{s,\mathrm{reference}}|.
\end{equation}
Positive $A_s$ means a larger absolute pooled gap under the specified observation process. This is a counterfactual contrast \emph{inside the simulation}, conditional on the archive and assumptions. It is not a causal effect of actual policing. All intervals describe simulation or specified resampling variation; none quantify all uncertainty about real crime or real deployments. We report effects and intervals without a family of unadjusted significance tests.

\subsection{Exploratory tract-level associations}
Actual simulated detected and total incident counts are retained for each tract, month, year, seed and condition. For the generative policy's detected and mixed streams, we fit a separate descriptive fractional-binomial logit model in each city. Its response is the mean seed-level tract-month detection fraction. Weights are the number of archived incident records in that tract-month, not twenty times that count. Predictors are Black share per 10 pp, log median household income divided by 10000, poverty share per 10 pp, and log one plus recorded-incident density per square kilometer. Month indicators and, for Baltimore, year indicators are included. Income coefficients are estimated separately by city, avoiding direct pooling of 2019 and 2022 nominal dollar values.

We resample complete seeds and spatial blocks, retaining all months and years within a selected block. Two hundred fifty bootstrap fits at each of 1.5, 3 and 6 km block widths assess interval sensitivity; 3 km is the main specification. This is a descriptive spatial sensitivity analysis, not a proof that residual dependence has been removed. A six-nearest-neighbor residual Moran statistic and model diagnostics are retained. The analysis does not assign area-category averages to tracts, does not treat administrative records as independent evidence about individual race, and makes no causal interpretation of coefficients.

\subsection{Synthetic replacement and robustness experiments}
For Baltimore 2019, a separate rolling-reference experiment compares raw previous-month training, simple balanced replacement and conditional tabular generation. Each condition trains the patrol generator on the same number of records and evaluates the same next-month incidents. Replacement retains approximately 70\% of previous-month real records and replaces the remainder with equal target numbers from the three known area categories, allowing at most one record difference between categories. The simple intervention samples real previous-month coordinates within each category. It provides a comparator for whether synthetic generation adds value beyond reweighting.

The conditional tabular model is fitted on at most 600 real records from each known area group using $X$, $Y$ and area category, with 200 epochs, batch size 500, packing factor 10, and equal category selection during training. Conditional proposals must both carry the requested category and fall in a tract with that actual geographic category. We allow at most twenty proposal batches per target group. A failure to meet the target is recorded as infeasible; it is not filled with mislabeled samples or selectively dropped without disclosure. Only complete 11-month seed trajectories support annual intervention comparisons. Partial feasible months are retained for diagnostic inspection, without being promoted to an annual mitigation estimate. This reference-stream experiment does not evaluate mitigation under endogenous feedback.

Exploratory sensitivity checks use paired seeds 100--104 in Baltimore 2019. For the generative and empirical policies, they vary radius (400, 1000 and 1500 feet), officer count (30, 90 and 120), and per-officer detection probability (0.5 and 0.95) in the detected stream; reporting probabilities 0.3 and 0.8 are evaluated in the mixed stream. A separate generative-policy check uses 1000 updates in all three streams. Every contrast uses the same five baseline seeds. These checks are limited robustness descriptions, not a global parameter-importance analysis.

\section{Results and discussion}
\subsection{Audit of the original quantitative claims}
The archived output confirms that the original 264 simulation rows represent 88 month-mode cells times three area categories, arising from eight annual sequences. The stored neighborhood regression table contains 279 rows but only three distinct outcome values, assigned from area-category averages. Those correlations and OLS estimates are excluded from this revision because their outcome construction cannot provide independent neighborhood-level evidence.

Baltimore 2019's original detected stream has zero White-category detections in eight months. Pooling its saved counts gives 800 detected records out of 32939 Black-category records and 53 out of 7924 White-category records: rates of 2.43\% and 0.67\%, respectively, and a pooled ratio of 3.63. This is an \emph{audit of legacy counts under legacy labels}, not a corrected simulation estimate. It also does not estimate the same quantity as the reported mean monthly epsilon-stabilized ratio of approximately 15714. The contrast illustrates how aggregation and the treatment of zeros can dominate a headline statistic. The main revised results below use regenerated outcomes, official tract covariates, corrected distances and consistent estimands.

\subsection{Rebuilt inputs and main comparisons}
The fixed municipal masks retain 145142 Baltimore records and 234073 Chicago records, including January initialization records. They exclude 36 and 809 archived records, respectively. Unknown demographic categories are retained and reported separately (Table~\ref{tab:data}). These counts supersede the conflicting manuscript and README counts; they are not estimates of all city crime.

\begin{figure}[H]\centering\singlespacing\includegraphics[width=\linewidth]{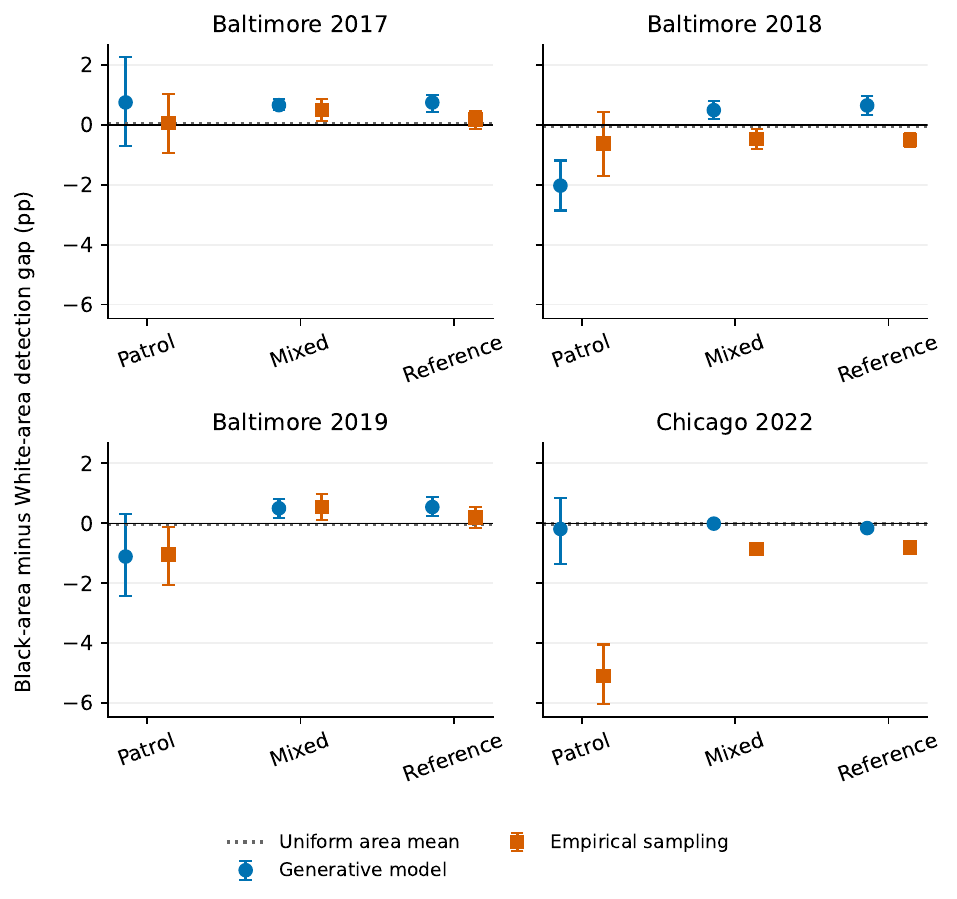}\caption{\textbf{Detection gaps across allocation and observation conditions.} Mean pooled detection gap in each city-year; corresponding coverage estimates appear in Table~\ref{tab:main}. Points compare the generative and empirical policies under patrol-detected, mixed, and full-record reference training. Error bars are 95\% whole-trajectory bootstrap intervals over 20 seeds. Dotted horizontal lines show uniform-area means. The gap denominator is recorded incidents in each area category, not residents.}\label{fig:main}\end{figure}

\begin{figure}[H]\centering\singlespacing\includegraphics[width=\linewidth]{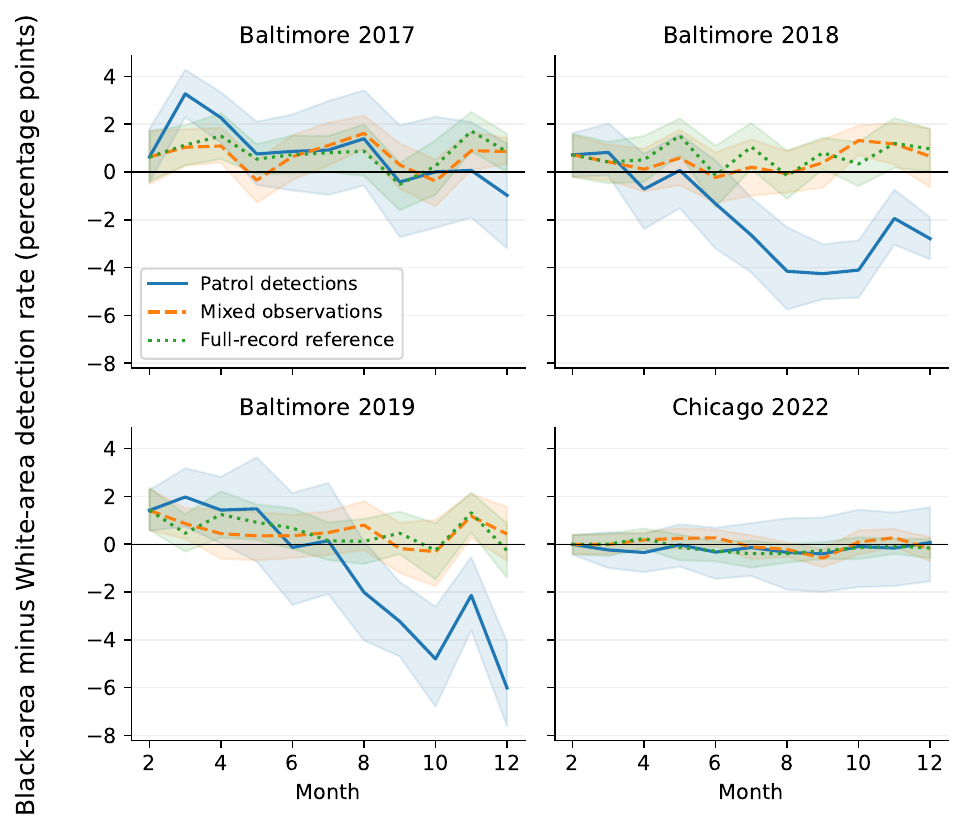}\caption{\textbf{Monthly detection gaps for the generative policy.} Monthly generative-policy detection gaps. Lines are means across 20 trajectories, with pointwise 95\% whole-trajectory bootstrap bands. The trajectories share a January initialization but evolve separately by observation stream. A positive value indicates greater modeled detection of recorded incidents in Black-majority tracts.}\label{fig:monthly}\end{figure}

Under detection-only training, generative-policy mean pooled gaps are Baltimore 2017: 0.75 [-0.69, 2.26] pp; Baltimore 2018: -2.03 [-2.86, -1.19] pp; Baltimore 2019: -1.11 [-2.42, 0.31] pp; Chicago 2022: -0.19 [-1.35, 0.83] pp (95\% simulation bootstrap intervals). The signed means do not describe the full range of seed-level disparity directions. In the full-record reference comparisons, empirical sampling has greater mean total detection coverage than the generative prototype in every city-year. This does not make coverage a sufficient fairness criterion. Table~\ref{tab:main} reports pooled gap and total coverage by policy and stream. Figure~\ref{fig:main} displays the mean comparisons and bootstrap intervals; Figure~\ref{fig:monthly} shows monthly generative-model gap trajectories. Monthly bands are pointwise and do not establish simultaneous evidence across all months. 7 of the 28 condition means have gap MCSE above the prespecified reporting flag of 0.5 pp. Thus the 20-seed budget does not provide uniformly precise signed-gap estimates. No direction of disparity is inferred solely from one seed, an isolated month, or an extreme finite replacement for a zero denominator.

\subsection{Observation feedback and spatial fit}
For the generative policy, detected-input feedback increases mean absolute pooled gap relative to the reference by 1.53--2.11 pp across the four city-years. Each corresponding 95\% paired bootstrap interval lies above zero. Mixed-input absolute-gap contrasts range from -0.23 to 0.09 pp, and their intervals include zero in each city-year. This supports a distinction between the two modeled observation streams under the chosen assumptions, rather than a universal assertion about citizen reporting or policing. Table~\ref{tab:feedback} reports paired absolute-gap and coverage contrasts against the full-record reference. These differences hold the archive, policy class and nominal resources fixed. They do not separate every mechanism of a real policing system, and the reference already inherits the archive's original recording biases. The mixed stream also embeds the assumption that citizen reporting is spatially uniform and independent of detection before their union is formed.

Under the full-record reference, the generative prototype's mean next-month spatial divergence is 0.121--0.159, compared with 0.023--0.040 for empirical sampling; the empirical mean is lower in every city-year. Its superior fit in these comparisons limits any claim that adversarial generation adds predictive value at the tested budget. Spatial fit is measured by Jensen--Shannon divergence on a fixed $12\times12$ city bounding grid, comparing 512 sampled positions with next-month recorded locations. It is a distributional diagnostic rather than a calibrated forecast score or an estimate of predictive accuracy for unrecorded crime. Rejection fractions and training losses are retained to distinguish a disparity estimate from a successfully fitted spatial model. The existence of a disparity under a generative allocator does not establish that adversarial generation is necessary to produce it.

\subsection{Synthetic replacement feasibility and outcomes}
\begin{figure}[H]\centering\singlespacing\includegraphics[width=\linewidth]{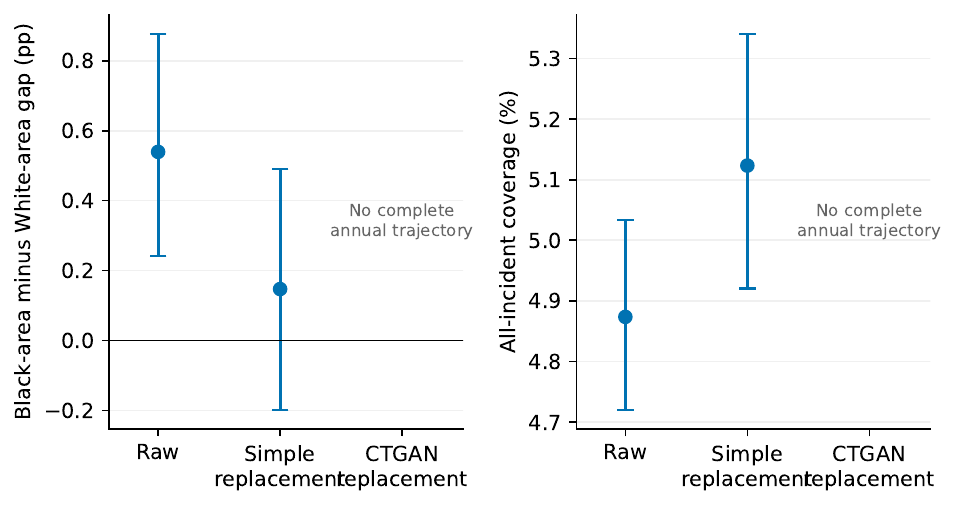}\caption{\textbf{Replacement outcomes and incomplete synthetic-generation trajectories.} Rolling-reference replacement experiment in Baltimore 2019. Annual outcomes are plotted only for complete 11-month seed trajectories. Conditional-generative synthesis failures are recorded in Table~\ref{tab:mitigation}; missing annual estimates are not replaced by results from selected feasible months. Error bars describe trajectory bootstrap variation.}\label{fig:mitigation}\end{figure}

Simple replacement yields a mean gap of 0.15 [-0.20, 0.49] pp and coverage of 5.12 [4.92, 5.34]\%, compared with raw-training means of 0.54 pp and 4.87\%. Its paired absolute-gap change is -0.04 pp [-0.38, 0.27], providing no clear reduction in disparity magnitude under this budget. Conditional-generative replacement meets all target counts in 113 of 220 seed-month fits and fails in 107; 0 of 20 seed trajectories are complete. No annual CTGAN mitigation effect is therefore estimated. Across the monthly fits, mean unfiltered proposal-label/geography agreement is 29.5\%, and the mean accepted-proposal fraction is 6.6\%. These diagnostics concern the specified proposal generator and acceptance rule. Table~\ref{tab:mitigation} separates feasibility from annual outcome estimation. Figure~\ref{fig:mitigation} shows only estimable complete-trajectory outcomes. Failed synthesis is a substantive result about this particular training and sampling specification; it does not show that every conditional generative model is incapable of geographic balancing. Changing the proposal budget or fitting a better spatial generator could change feasibility and would require a separately documented experiment.

Even geographically consistent balancing would not establish fairness in individual police contact. The intervention balances replacement samples across area categories, not actual victim or offender identities, exposure to coercion, or access to services. Any reduction in absolute detection differences must also be considered alongside changes in total detection coverage. We therefore withdraw the earlier claim that the synthetic experiment demonstrates a general need for accompanying policy intervention; that conclusion is outside what this simulation identifies.

\subsection{Tract-level associations and robustness}
The tract-specific models use 6456 Baltimore tract-month cells spanning 196 tracts and 8703 Chicago tract-month cells spanning 817 tracts. These are repeated geographic observations, not independent people or independent annual neighborhoods. At 3 km block width, all four intervals for the Black-share odds ratio include 1. Residual spatial association remains substantial (Moran statistics 0.63--0.67); the regression is therefore retained as an exploratory description, not independent confirmation that demographics explain deployment disparity. Table~\ref{tab:regression} reports the Black-share association under the 3 km block specification. Full coefficients, alternate block sizes, convergence checks and residual diagnostics accompany the supplement. These associations can arise from geographic incident concentration and the model's exposure mechanism. They do not demonstrate that a person's race caused detection, or that changing an area's racial composition would change actual police behavior.

Using 1000 instead of 200 generative-model updates changes the mean signed pooled gap in the five paired Baltimore 2019 seeds as follows: detected: -1.30 pp; mixed: -0.05 pp; reference: +0.03 pp. The full parameter contrasts are supplied in the supplement, including physical detection and reporting changes. These results are specific to the five-seed robustness subset and should not be substituted for the 20-seed main estimates. In particular, changes under a larger training budget are evidence of dependence on model specification. A parameter sweep with five seeds cannot rank the global importance of officer count, radius and reporting across their entire plausible ranges. Nor can an assumed uniform reporting mechanism validate the empirical pattern of calls for service.

\subsection{Interpretation and limitations}
The defensible contribution is a reproducible spatial simulation audit: investigators can compare allocation and observation assumptions while tracing every outcome to recorded input locations and saved counts. Treating the observation stream as an explicit component helps distinguish feedback contrasts from ordinary temporal variation. Reporting absolute gaps alongside coverage prevents an apparently favorable ratio from concealing very low detection in every group. Geographic validation of synthetic records also prevents a categorical label from being mistaken for a verified spatial demographic attribute.

Several limits materially constrain interpretation. First, the archive contains recorded incidents, not latent crime. Its original extraction cannot be fully reconstructed from the supplied materials; hashes freeze the analysis inputs but do not independently certify their history or completeness. Second, the cities differ in offense universe and calendar period. Three Baltimore annual sequences and one Chicago year do not provide a balanced design for attributing differences to city or historical change. Third, tract demographics are ecological proxies, and Other combines heterogeneous populations. The ACS estimates and majority threshold add measurement uncertainty beyond the plotted simulation intervals.

Fourth, fixed monthly point deployments and a homogeneous Noisy-OR probability omit patrol movement, temporal availability, enforcement discretion, victim reporting differences, and institutional decision-making. January full observation and the carried-forward sparse-data rule are explicit modeling choices. The municipal masks can include uninhabited or nonresidential areas, and archived geocodes may be approximate. Fifth, the generative architecture is a prototype with a fixed update budget, not an optimized location-prediction system. Sampling validity, convergence and spatial fidelity must be assessed before assigning sociological meaning to its outputs.

Finally, twenty seeds characterize only a limited distribution of stochastic model realizations under these inputs. Bootstrap intervals, alternative spatial block sizes and five-seed sensitivity checks do not account for every uncertainty in policing or ensure transportability. This work does not estimate deterrence, wrongful police contact, protection, arrest, or the effectiveness of a real intervention. The ethical interpretation of higher or lower detection requires community and institutional context that the numerical model does not observe~\cite{selbst2019fairness}.

\section{Conclusions}
We revised the framework to use explicit temporal observation streams, physical spatial distances, geographically sourced area labels, consistent count-based metrics, simpler allocation baselines, repeated-seed uncertainty and tract-specific outcomes. The corrected study evaluates simulated geographic exposure conditional on a frozen incident archive. It does not retain unsupported claims about individual racial detection, universal feedback direction, or validated conditional-generative mitigation. A useful audit must make synthesis failures, zero denominators, model fit, coverage and remaining uncertainty visible alongside disparity summaries. The accompanying additional files supply the inputs, protocols, diagnostics and executable analyses needed to inspect and reproduce the results.

\section*{List of abbreviations}
ACS: American Community Survey; CTGAN: conditional tabular generative adversarial network; GAN: generative adversarial network; pp: percentage points; MCSE: Monte Carlo standard error.

\section*{Competing interests}
The authors declare that they have no known competing financial interests or personal relationships that could have influenced the work reported in this paper.

\section*{Authors' contributions}
\textbf{Pronob Kumar Barman:} Conceptualization; methodology; software; formal analysis; investigation; visualization; writing---original draft; writing---review and editing.

\textbf{Pronoy Kumar Barman:} Methodology; data curation; validation; formal analysis; writing---review and editing.

\section*{Authors' information}
Pronob Kumar Barman: Department of Information Systems, University of Maryland, Baltimore County (UMBC), Baltimore, Maryland, USA. Pronoy Kumar Barman: Department of Statistics, Jagannath University, Dhaka, Bangladesh. Correspondence: \href{mailto:pbarman1@umbc.edu}{pbarman1@umbc.edu}.

\section*{Acknowledgements}
The authors acknowledge the Baltimore and Chicago public-data providers and the U.S. Census Bureau for making the underlying data sources available.

\section*{Funding}
This research received no external funding.

\section*{Ethics approval and consent}
The study uses secondary publicly available crime incident records from Baltimore and Chicago and demographic information from the U.S. Census Bureau. No direct participant recruitment or intervention was conducted. 

\section*{Data, code and preprint availability}
An earlier version of this manuscript is publicly available on arXiv~\cite{barmanPreprint}. The accompanying replication files contain frozen incident inputs, official demographic extracts and geography, source hashes, the experiment protocol, code, environment versions and saved outputs. The earlier public replication repository~\cite{githubRepository} documents a historical implementation; its contents should not be assumed to match the revised results. Input provenance and the limits of reconstructing the original raw extraction are described in Methods. The revised manuscript, analysis code, frozen replication files and saved outputs are available at \url{https://github.com/pronob29/predictive-policing-spatial-audit-jdmdh}. The additional files identify the frozen analysis used for this article. The repository is versioned separately from the earlier implementation.

\section*{Declaration of AI-assisted work}
An AI language assistant (OpenAI Codex) was used only for writing assistance, namely language editing and improving the clarity and readability of the text. It was not used to generate research ideas, design the study, write or audit code, conduct analyses, or produce results or figures. The research concept, design, implementation, analysis, results and conclusions are the authors' own work, and the authors take full responsibility for the content of this article.

\section*{Additional files}
\begin{description}
\item[Additional file 1] \texttt{Additional\_file\_1\_supplement.pdf} (PDF): supplementary methods, diagnostics, secondary ratios and sensitivity analyses.
\item[Additional file 2] \texttt{Additional\_file\_2\_result\_tables.zip} (ZIP): machine-readable result tables, protocol, validation records and analysis metadata.
\item[Additional file 3] \texttt{Additional\_file\_3\_replication\_01.zip} (ZIP): a numbered volume of the frozen replication inputs, code, saved trajectories and source records.
\item[Additional file 4] \texttt{Additional\_file\_4\_replication\_02.zip} (ZIP): a numbered volume of the frozen replication inputs, code, saved trajectories and source records.
\item[Additional file 5] \texttt{Additional\_file\_5\_replication\_03.zip} (ZIP): a numbered volume of the frozen replication inputs, code, saved trajectories and source records.
\item[Additional file 6] \texttt{Additional\_file\_6\_replication\_04.zip} (ZIP): a numbered volume of the frozen replication inputs, code, saved trajectories and source records.
\item[Additional file 7] \texttt{Additional\_file\_7\_replication\_05.zip} (ZIP): a numbered volume of the frozen replication inputs, code, saved trajectories and source records.
\item[Additional file 8] \texttt{Additional\_file\_8\_replication\_06.zip} (ZIP): a numbered volume of the frozen replication inputs, code, saved trajectories and source records.
\item[Additional file 9] \texttt{Additional\_file\_9\_replication\_07.zip} (ZIP): a numbered volume of the frozen replication inputs, code, saved trajectories and source records.
\item[Additional file 10] \texttt{Additional\_file\_10\_replication\_08.zip} (ZIP): a numbered volume of the frozen replication inputs, code, saved trajectories and source records.
\end{description}
Extract all replication ZIP volumes into the same directory. The complete file inventory is supplied in \texttt{replication\_volume\_manifest.json}. The tract-month count table is regenerated from the included run records. Separate figure files and editable manuscript sources accompany these materials.

\clearpage
\section*{Tables}
\begin{table}[H]\centering\singlespacing
\small
\caption{\textbf{Archived and retained incident records.} Retained counts include January initialization. Unknown counts cover the full retained year; they are included in total coverage but excluded from Black/White rate denominators.}\label{tab:data}
\begin{tabular}{|l|l|l|l|l|l|}\hline
City & Year & Archived & Retained & Feb.--Dec. & Unknown \\ \hline
Baltimore & 2017 & 51511 & 51498 & 46950 & 330 \\ \hline
Baltimore & 2018 & 47856 & 47843 & 44129 & 328 \\ \hline
Baltimore & 2019 & 45811 & 45801 & 42133 & 269 \\ \hline
Chicago & 2022 & 234882 & 234073 & 218814 & 1433 \\
\hline\end{tabular}\end{table}

\begin{table}[H]\centering\singlespacing
\small
\caption{\textbf{Main simulation outcomes by city, policy, and observation stream.} Main results: means and 95\% whole-trajectory bootstrap intervals over 20 seeds. Gap is Black-area minus White-area detection in pp; coverage is detected records as a percentage of all archived analysis incidents. Conditions are paired by seed.}\label{tab:main}
\begin{tabular}{|l|l|l|l|l|}\hline
City-year & Policy & Training stream & Gap (pp) & Coverage (\%) \\ \hline
Balt. 2017 & Generative & Detected & 0.75 [-0.69, 2.26] & 6.41 [6.12, 6.69] \\ \hline
Balt. 2017 & Generative & Mixed & 0.66 [0.49, 0.86] & 4.70 [4.52, 4.88] \\ \hline
Balt. 2017 & Generative & Reference & 0.75 [0.44, 1.01] & 4.73 [4.60, 4.88] \\ \hline
Balt. 2017 & Empirical & Detected & 0.06 [-0.94, 1.03] & 8.79 [8.56, 8.99] \\ \hline
Balt. 2017 & Empirical & Mixed & 0.50 [0.12, 0.86] & 8.50 [8.42, 8.58] \\ \hline
Balt. 2017 & Empirical & Reference & 0.18 [-0.12, 0.48] & 8.26 [8.17, 8.35] \\ \hline
Balt. 2017 & Uniform & Reference & 0.05 [-0.19, 0.27] & 2.97 [2.90, 3.04] \\ \hline
Balt. 2018 & Generative & Detected & -2.03 [-2.86, -1.19] & 7.43 [7.18, 7.67] \\ \hline
Balt. 2018 & Generative & Mixed & 0.49 [0.20, 0.79] & 5.01 [4.82, 5.18] \\ \hline
Balt. 2018 & Generative & Reference & 0.65 [0.33, 0.97] & 5.10 [4.90, 5.30] \\ \hline
Balt. 2018 & Empirical & Detected & -0.62 [-1.70, 0.44] & 8.74 [8.36, 9.06] \\ \hline
Balt. 2018 & Empirical & Mixed & -0.47 [-0.81, -0.12] & 9.07 [8.91, 9.22] \\ \hline
Balt. 2018 & Empirical & Reference & -0.50 [-0.74, -0.25] & 8.67 [8.55, 8.79] \\ \hline
Balt. 2018 & Uniform & Reference & -0.04 [-0.23, 0.13] & 3.01 [2.93, 3.10] \\ \hline
Balt. 2019 & Generative & Detected & -1.11 [-2.42, 0.31] & 7.50 [7.25, 7.74] \\ \hline
Balt. 2019 & Generative & Mixed & 0.50 [0.19, 0.81] & 5.05 [4.87, 5.22] \\ \hline
Balt. 2019 & Generative & Reference & 0.54 [0.24, 0.86] & 4.87 [4.71, 5.03] \\ \hline
Balt. 2019 & Empirical & Detected & -1.05 [-2.05, -0.11] & 9.14 [8.83, 9.41] \\ \hline
Balt. 2019 & Empirical & Mixed & 0.55 [0.11, 0.98] & 9.19 [9.08, 9.30] \\ \hline
Balt. 2019 & Empirical & Reference & 0.20 [-0.17, 0.54] & 8.93 [8.81, 9.04] \\ \hline
Balt. 2019 & Uniform & Reference & -0.05 [-0.29, 0.15] & 3.02 [2.92, 3.15] \\ \hline
Chic. 2022 & Generative & Detected & -0.19 [-1.35, 0.83] & 1.78 [1.66, 1.93] \\ \hline
Chic. 2022 & Generative & Mixed & -0.01 [-0.12, 0.09] & 1.46 [1.44, 1.49] \\ \hline
Chic. 2022 & Generative & Reference & -0.16 [-0.27, -0.05] & 1.55 [1.52, 1.59] \\ \hline
Chic. 2022 & Empirical & Detected & -5.10 [-6.05, -4.05] & 3.76 [3.56, 3.93] \\ \hline
Chic. 2022 & Empirical & Mixed & -0.86 [-1.04, -0.67] & 3.24 [3.19, 3.29] \\ \hline
Chic. 2022 & Empirical & Reference & -0.81 [-1.01, -0.59] & 3.16 [3.10, 3.20] \\ \hline
Chic. 2022 & Uniform & Reference & -0.03 [-0.12, 0.05] & 1.20 [1.18, 1.23] \\
\hline\end{tabular}\end{table}

\begin{table}[H]\centering\singlespacing
\small
\caption{\textbf{Paired feedback contrasts for the generative policy.} Generative-policy feedback contrasts against the same policy's full-record reference. Values are mean paired differences and 95\% whole-pair bootstrap intervals. Absolute gap uses each seed's magnitude before averaging.}\label{tab:feedback}
\begin{tabular}{|l|l|l|l|}\hline
City-year & Training stream & Absolute-gap change (pp) & Coverage change (pp) \\ \hline
Balt. 2017 & Detected & 2.02 [1.12, 2.96] & 1.67 [1.42, 1.94] \\ \hline
Balt. 2017 & Mixed & -0.23 [-0.49, 0.03] & -0.03 [-0.23, 0.17] \\ \hline
Balt. 2018 & Detected & 1.53 [0.72, 2.38] & 2.34 [1.98, 2.68] \\ \hline
Balt. 2018 & Mixed & -0.08 [-0.44, 0.30] & -0.09 [-0.36, 0.17] \\ \hline
Balt. 2019 & Detected & 2.11 [1.25, 3.09] & 2.63 [2.34, 2.90] \\ \hline
Balt. 2019 & Mixed & 0.09 [-0.26, 0.39] & 0.18 [-0.03, 0.39] \\ \hline
Chic. 2022 & Detected & 1.63 [0.98, 2.37] & 0.23 [0.11, 0.36] \\ \hline
Chic. 2022 & Mixed & -0.03 [-0.12, 0.05] & -0.09 [-0.12, -0.06] \\
\hline\end{tabular}\end{table}

\begin{table}[H]\centering\singlespacing
\small
\caption{\textbf{Synthetic-replacement feasibility and outcomes.} Baltimore 2019 rolling-reference replacement experiment. Annual estimates require complete 11-month trajectories. Failure to synthesize valid samples is not imputed or replaced by a selected-month estimate.}\label{tab:mitigation}
\begin{tabular}{|l|l|l|l|l|}\hline
Condition & Feasible fits & Complete seeds & Gap (pp) & Coverage (\%) \\ \hline
Raw & 220/220 & 20/20 & 0.54 [0.24, 0.88] & 4.87 [4.72, 5.03] \\ \hline
Simple replacement & 220/220 & 20/20 & 0.15 [-0.20, 0.49] & 5.12 [4.92, 5.34] \\ \hline
CTGAN replacement & 113/220 & 0/20 & Not estimable & Not estimable \\
\hline\end{tabular}\end{table}

\begin{table}[H]\centering\singlespacing
\small
\caption{\textbf{Exploratory tract-level demographic associations.} Exploratory Black-share association per 10 pp in actual tract-month simulated detection. Odds ratios and 95\% seed/spatial-block resampling intervals use 3 km blocks, 250 resamples, and covariate-adjusted fractional-binomial models. These are not individual-race or causal effects.}\label{tab:regression}
\begin{tabular}{|l|l|l|l|l|}\hline
City & Training stream & Odds ratio [interval] & Tracts & Tract-months \\ \hline
Baltimore & Detected & 0.866 [0.773, 1.014] & 196 & 6456 \\ \hline
Baltimore & Mixed & 0.973 [0.938, 1.038] & 196 & 6456 \\ \hline
Chicago & Detected & 0.999 [0.941, 1.071] & 817 & 8703 \\ \hline
Chicago & Mixed & 0.997 [0.969, 1.025] & 817 & 8703 \\
\hline\end{tabular}\end{table}

\clearpage
\section*{Figures}
\begin{figure}[H]\centering\singlespacing\includegraphics[width=\linewidth]{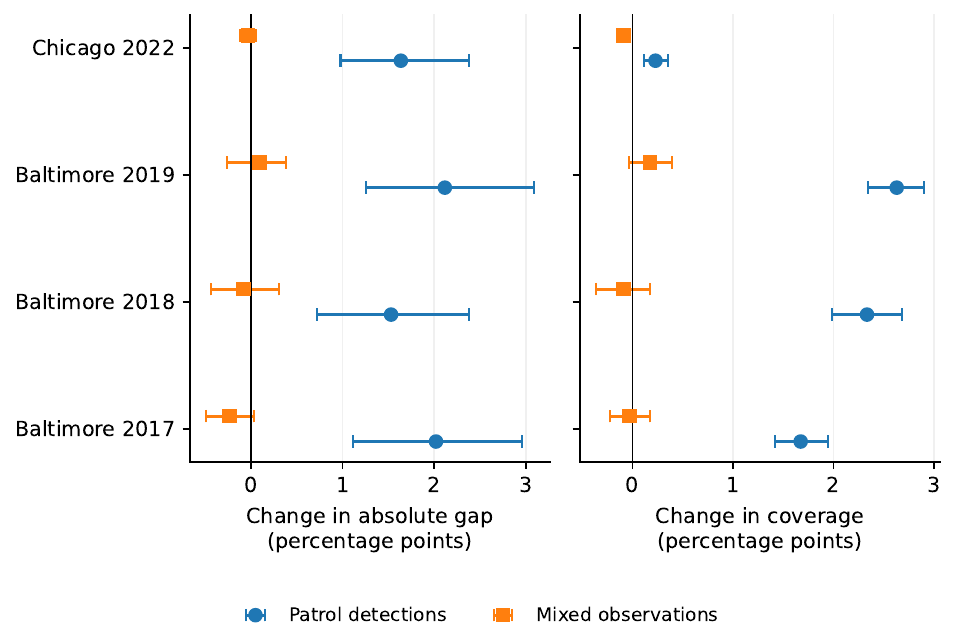}\caption{\textbf{Feedback contrasts in disparity magnitude and coverage.} Paired differences between observed-input generative trajectories and their full-record references. Positive absolute-gap changes mean greater disparity magnitude within this simulation. Coverage differences are percentage points. Error bars resample whole paired trajectories; these are not real-world causal effects.}\label{fig:feedback}\end{figure}

\begin{thebibliography}{99}
\begin{spacing}{1.15}
\bibitem{ensign2018runaway} Ensign D, Friedler SA, Neville S, Scheidegger C, Venkatasubramanian S: Runaway feedback loops in predictive policing. In \emph{Proceedings of the 1st Conference on Fairness, Accountability and Transparency}. \emph{Proceedings of Machine Learning Research} 2018, \textbf{81}:160--171. \url{https://proceedings.mlr.press/v81/ensign18a.html}.
\bibitem{lum2016predict} Lum K, Isaac W: To predict and serve? \emph{Significance} 2016, \textbf{13}(5):14--19. \url{https://doi.org/10.1111/j.1740-9713.2016.00960.x}.
\bibitem{richardson2019dirty} Richardson R, Schultz JM, Crawford K: Dirty data, bad predictions: How civil rights violations impact police data, predictive policing systems, and justice. \emph{New York University Law Review Online} 2019, \textbf{94}:192--233. \url{https://www.nyulawreview.org/wp-content/uploads/2019/04/NYULawReview-94-Richardson-Schultz-Crawford.pdf}.
\bibitem{wu2024fairness} Wu J, Frias-Martinez V: Improving the fairness of deep-learning, short-term crime prediction with under-reporting-aware models. \emph{arXiv preprint} 2024, arXiv:2406.04382v2. \url{https://arxiv.org/abs/2406.04382v2}.
\bibitem{semsar2026comparative} Semsar S, Prabhu KL, Waters G, Foulds J: A comparative simulation study of the fairness and accuracy of predictive policing systems in Baltimore City. \emph{arXiv preprint} 2026, arXiv:2602.02566. \url{https://arxiv.org/abs/2602.02566}.
\bibitem{goodfellow2014generative} Goodfellow I, Pouget-Abadie J, Mirza M, Xu B, Warde-Farley D, Ozair S, Courville A, Bengio Y: Generative adversarial nets. In \emph{Advances in Neural Information Processing Systems} 2014, \textbf{27}:2672--2680. \url{https://arxiv.org/abs/1406.2661}.
\bibitem{xu2019modeling} Xu L, Skoularidou M, Cuesta-Infante A, Veeramachaneni K: Modeling tabular data using conditional GAN. In \emph{Advances in Neural Information Processing Systems} 2019, \textbf{32}:7333--7343. \url{https://papers.neurips.cc/paper_files/paper/2019/hash/254ed7d2de3b23ab10936522dd547b78-Abstract.html}.
\bibitem{selbst2019fairness} Selbst AD, Boyd D, Friedler SA, Venkatasubramanian S, Vertesi J: Fairness and abstraction in sociotechnical systems. In \emph{Proceedings of the Conference on Fairness, Accountability, and Transparency}. New York: ACM; 2019:59--68. \url{https://doi.org/10.1145/3287560.3287598}.
\bibitem{mohler2015randomized} Mohler GO, Short MB, Malinowski S, Johnson M, Tita GE, Bertozzi AL, Brantingham PJ: Randomized controlled field trials of predictive policing. \emph{Journal of the American Statistical Association} 2015, \textbf{110}(512):1399--1411. \url{https://doi.org/10.1080/01621459.2015.1077710}.
\bibitem{ferguson2017policing} Ferguson AG: Policing predictive policing. \emph{Washington University Law Review} 2017, \textbf{94}(5):1109--1189. \url{https://journals.library.wustl.edu/lawreview/article/id/3851/}.
\bibitem{mehrabi2021survey} Mehrabi N, Morstatter F, Saxena N, Lerman K, Galstyan A: A survey on bias and fairness in machine learning. \emph{ACM Computing Surveys} 2021, \textbf{54}(6):115:1--115:35. \url{https://doi.org/10.1145/3457607}.
\bibitem{berk2021fairness} Berk R, Heidari H, Jabbari S, Kearns M, Roth A: Fairness in criminal justice risk assessments: The state of the art. \emph{Sociological Methods \& Research} 2021, \textbf{50}(1):3--44. \url{https://doi.org/10.1177/0049124118782533}.
\bibitem{chouldechova2017fair} Chouldechova A: Fair prediction with disparate impact: A study of bias in recidivism prediction instruments. \emph{Big Data} 2017, \textbf{5}(2):153--163. \url{https://doi.org/10.1089/big.2016.0047}.
\bibitem{dressel2018accuracy} Dressel J, Farid H: The accuracy, fairness, and limits of predicting recidivism. \emph{Science Advances} 2018, \textbf{4}(1):eaao5580. \url{https://doi.org/10.1126/sciadv.aao5580}.
\bibitem{wang2023interpretable} Wang C, Han B, Patel B, Rudin C: In pursuit of interpretable, fair and accurate machine learning for criminal recidivism prediction. \emph{Journal of Quantitative Criminology} 2023, \textbf{39}(2):519--581. \url{https://doi.org/10.1007/s10940-022-09545-w}.
\bibitem{zhang2018fairness} Zhang J, Bareinboim E: Fairness in decision-making---the causal explanation formula. In \emph{Proceedings of the AAAI Conference on Artificial Intelligence} 2018, \textbf{32}(1). \url{https://doi.org/10.1609/aaai.v32i1.11564}.
\bibitem{kim2021counterfactual} Kim H, Shin S, Jang J, Song K, Joo W, Kang W, Moon IC: Counterfactual fairness with disentangled causal effect variational autoencoder. In \emph{Proceedings of the AAAI Conference on Artificial Intelligence} 2021, \textbf{35}(9):8128--8136. \url{https://doi.org/10.1609/aaai.v35i9.16990}.
\bibitem{almasoud2025bias} Almasoud AS, Idowu JA: Algorithmic fairness in predictive policing. \emph{AI and Ethics} 2025, \textbf{5}:2323--2337. \url{https://doi.org/10.1007/s43681-024-00541-3}.
\bibitem{hung2023ethics} Hung TW, Yen CP: Predictive policing and algorithmic fairness. \emph{Synthese} 2023, \textbf{201}:206. \url{https://doi.org/10.1007/s11229-023-04189-0}.
\bibitem{ziosi2024contested} Ziosi M, Pruss D: Evidence of what, for whom? The socially contested role of algorithmic bias in a predictive policing tool. In \emph{Proceedings of the 2024 ACM Conference on Fairness, Accountability, and Transparency}. New York: ACM; 2024:1596--1608. \url{https://arxiv.org/abs/2405.07715}.
\bibitem{BPD2019} Baltimore City Police Department: Part 1 Crime Data (Legacy SRS), dataset metadata. \url{https://www.arcgis.com/home/item.html?id=e0992dddbbf64231976d5d57763ec4f5}. Accessed 22 October 2026. Incident analysis uses the frozen processed archive described in Methods.
\bibitem{CPD2022} City of Chicago: Crimes---2001 to present. Dataset ijzp-q8t2. \url{https://data.cityofchicago.org/d/ijzp-q8t2}. Accessed 22 October 2026. Incident analysis uses the frozen processed archive described in Methods.
\bibitem{chicagoBoundary} City of Chicago: Boundaries---City. Dataset qqq8-j68g. \url{https://data.cityofchicago.org/resource/qqq8-j68g.geojson}. Accessed 22 October 2026.
\bibitem{ACS2022} U.S. Census Bureau: American Community Survey five-year detailed tables B02001, B19013 and B17001, 2019 and 2022 releases. \url{https://www2.census.gov/programs-surveys/acs/summary_file/}. Exact file URLs and hashes are recorded in the accompanying source manifest. Accessed 22 October 2026.
\bibitem{censusGeography} U.S. Census Bureau: TIGERweb ACS2019 tract geography and TIGER/Line 2022 Illinois census tracts. \url{https://tigerweb.geo.census.gov/arcgis/rest/services/TIGERweb/tigerWMS_ACS2019/MapServer}; \url{https://www2.census.gov/geo/tiger/TIGER2022/TRACT/tl_2022_17_tract.zip}. Accessed 22 October 2026.
\bibitem{BJS2022} Thompson A, Tapp SN: \emph{Criminal Victimization, 2022}. Washington, DC: Bureau of Justice Statistics; 2023. NCJ 307089. \url{https://bjs.ojp.gov/document/cv22.pdf}.
\bibitem{barmanPreprint} Barman PK, Barman PK: Unmasking algorithmic bias in predictive policing: A GAN-based simulation framework with multi-city temporal analysis. \emph{arXiv preprint} 2026, arXiv:2603.18987v2. \url{https://arxiv.org/abs/2603.18987}.
\bibitem{githubRepository} Barman PK: Predictive-policing-bias-gan, historical replication repository. \url{https://github.com/pronob29/predictive-policing-bias-gan}. This historical implementation is distinct from the frozen revised analysis supplied with this manuscript.
\end{spacing}
\end{thebibliography}
\end{document}